\documentclass{article}

\usepackage{iclr2027_conference,times}

\usepackage[utf8]{inputenc}
\usepackage[T1]{fontenc}
\usepackage{microtype}
\usepackage{graphicx}
\usepackage{subcaption}
\usepackage{booktabs}
\usepackage{amsmath,amssymb,amsthm}
\usepackage{hyperref}
\hypersetup{colorlinks=true,linkcolor=blue!60!black,citecolor=blue!60!black,urlcolor=blue!60!black}
\usepackage{url}
\usepackage{xcolor}
\usepackage{tikz}
\usetikzlibrary{arrows.meta,positioning,fit,shapes,calc,backgrounds}
\usepackage{algorithm}
\usepackage{algpseudocode}
\usepackage{listings}
\usepackage{tabularx}
\usepackage{array}

\lstdefinestyle{prompt}{
  basicstyle=\ttfamily\footnotesize,
  breaklines=true,
  columns=fullflexible,
  keepspaces=true,
  frame=single,
  framesep=4pt,
  xleftmargin=4pt,
  xrightmargin=4pt,
  aboveskip=4pt,
  belowskip=4pt,
}

\title{Beyond Temporal Proximity: \\Explanatory World Models for \\Credit Assignment in Reinforcement Learning}

\author{Oriol Corcoll$^{1,*}$, Qiyao Wei$^{2}$, Ciar\'{a}n Mark Gilligan-Lee$^{1}$ \\
$^{1}$Spotify \quad $^{2}$University of Cambridge
}

\iclrfinalcopy

\begin{document}
\maketitle
\lhead{Preprint}
{\renewcommand{\thefootnote}{\fnsymbol{footnote}}%
 \footnotetext[1]{Corresponding author: \texttt{ocorcoll@gmail.com}}}

\begin{abstract}
Generative Recommenders increasingly use Reinforcement Learning (RL) to optimize outcomes that may appear hundreds or thousands of interactions after a decision.
Standard RL propagates reward according to temporal proximity rather than explicit relevance, attenuating the signal assigned to distant but important actions and creating a source of sample inefficiency.
We introduce Explanatory World Models (EWM), an attribution framework that uses explanations to identify the earlier events that plausibly contributed to a realized outcome. We instantiate EWM with a frontier language model and use its explanations as learning signals using a conservative hierarchical operator that recursively redistributes reward while preserving the undiscounted episode return.
To investigate what constitutes a useful explanation, we operationalize three complementary variants inspired by Halpern's actual causation, Lombrozo's explanatory simplicity, and Deutsch's hard-to-vary explanations.
We evaluate EWM on two controlled long-horizon RL benchmarks, a public sequential-recommendation benchmark, and an industry-scale generative recommender. 
On Crafter and WordCraft, EWM substantially outperforms PPO and LaRe, with its largest gains when rewards are most delayed. On KuaiRand, EWM outperforms SFT and LaRe offline and A2C online.
Finally, in an offline evaluation of an industry generative recommender, EWM considerably improves Recall and increases behavioral proxies.
These results suggest that explanations can provide structured and reusable credit for learning from delayed outcomes.
\end{abstract}

\section{Introduction}
\label{sec:intro}
In long-horizon decision-making problems, the consequences of an action may become apparent only after hundreds or thousands of subsequent decisions. An agent might gather a resource early in an episode but use it hundreds of steps later to complete an achievement spanning thousands of steps.
For example, a podcast recommendation surfaced today may introduce a listener to a show, shape their consumption over the following weeks, and ultimately affect whether they remain active months later. A recommender optimized to increase activity must determine which actions deserve credit.

This is the problem of temporal \emph{credit assignment}~\citep{minsky1961credit,sutton2018rl}. Standard RL algorithms propagate reward backward based on temporal proximity, i.e., assigning more credit to recent actions through discounted returns.
This mechanism is computationally convenient, but it embeds a temporal prior: actions closer to a reward are treated as more relevant. 
Given enough trajectories, an agent may learn to average out these spurious associations and recover the true relations between actions and reward. Doing so requires learning the structure of the environment almost entirely through trial and error. The length of the delay is therefore translated into sample inefficiency, where the further the action, the more interactions are required to identify its contribution.

A substantial body of work improves on raw discounted returns through return decomposition, hindsight information, or counterfactual reasoning ~\citep{arjonamedina2019rudder,patil2020aligned,harutyunyan2019hca,mesnard2021counterfactual}. These approaches demonstrate that credit does not need to be assigned solely according to the original reward timing.
Rather than requiring the policy's critic to discover relevance only through future-return prediction, we pursue a complementary direction motivated by work on explanations and causal reasoning.
Cognitive and social scientists treat explanations as central to human inquiry and cognition: explanations organize scientific understanding, drive theory formation, and shape learning, causal inference, and generalization~\citep{hempel1948explanation,gopnik1998explanation,keil2006explanation,lombrozo2006explanation, felekis2026causal, gilligan2020causing, reynaud2022d}.
Different accounts of explanation emphasize different criteria for identifying which preceding events matter~\citep{lombrozo2006explanation}. Halpern and Pearl ground explanation in actual causation~\citep{halpern2005explanations}; Lombrozo emphasizes explanatory simplicity~\citep{lombrozo2007simplicity}; and Deutsch argues that good explanations are hard to vary and possess reach beyond the phenomenon for which they were constructed~\citep{deutsch2011beginning}.

We present \textbf{Explanatory World Models (EWM)}, a framework that attributes an observed event/reward to the earlier events/actions that best explain it. Given a trajectory and an event within it, an EWM returns a weighted set of preceding events. 
A redistribution operator then transports the reward to those events, changing where the learning signal is applied while preserving the episode’s return. EWM therefore modifies neither the environment nor the policy class; it inserts an attribution step between observing a reward and updating the policy.
\begin{figure}[t]
    \centering
    \includegraphics[width=0.7\linewidth]{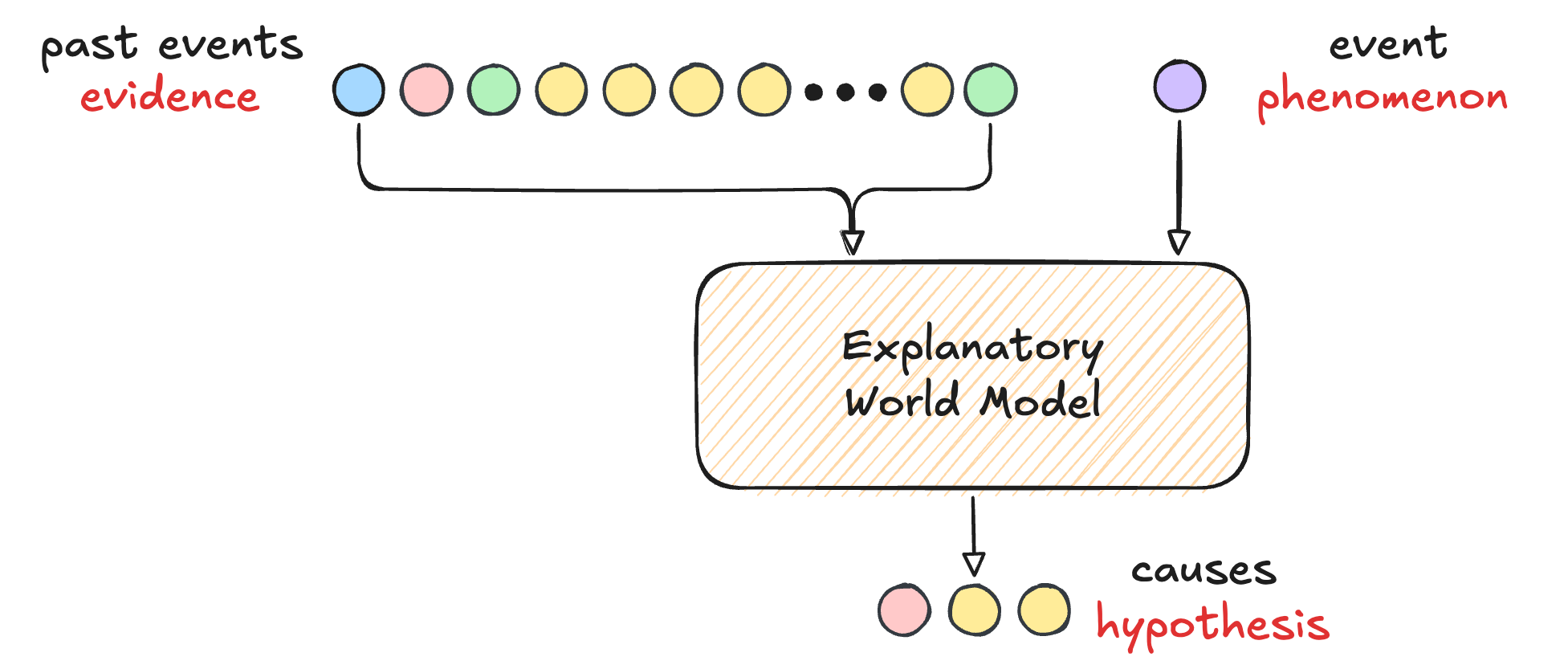}
    \caption{Explanatory World Model framework. EWMs take evidence (past events) and a phenomenon (the event to explain), and output a hypothesis/explanation about the events causing the phenomenon.}
    \label{fig:ewm}
\end{figure}

To convert explanations into a learning signal, we introduce a \emph{conservative hierarchical redistribution operator}. Explanations provide plausible causes to an outcome; instead of following a temporal bias, we can follow those relations to reallocate value to them.
Since EWM explains outcomes, not just rewards, we can apply this process recursively allowing to optimize value for extremely long horizons.
Moreover, a key benefit of explanation-based redistribution is its independence from the reward function. Changing the reward function therefore changes how an outcome is valued, but not which preceding events explain it, allowing explanations to be reused across reward functions.

We instantiate EWM with a frontier language model, using its prior knowledge to explain observed outcomes without having to learn the environment’s relations from interaction alone. 
We evaluate EWM in both long-horizon reasoning and real-world recommendation settings. On Crafter, EWM outperforms PPO across all eight tested combinations of discount factor and episode-length cap. On WordCraft, EWM outperforms PPO~\citep{schulman2017ppo} on the harder splits. In the offline KuaiRand setting, EWM outperforms SFT and LaRe. In the online KuaiRand setting, EWM outperforms PPO and achieves performance competitive with an oracle method.
In an offline replay evaluation of an industry-scale generative recommender, RLFT with EWM increases recall and independent measures of future engagement by \(~3\times\) over the SFT baseline, indicating that its gains extend beyond reproducing logged recommendations to downstream engagement proxies.

\paragraph{Contributions:}
\begin{itemize}
  \setlength{\topsep}{0pt}
  \setlength{\partopsep}{0pt}
  \setlength{\itemsep}{2pt}
  \setlength{\parskip}{0pt}
  \setlength{\parsep}{0pt}
  \item We introduce \textbf{Explanatory World Models}, a general framework that recasts long-horizon credit assignment as explanation-based attribution.
  \item We develop a \textbf{conservative hierarchical redistribution} operator that propagates reward backward through an explanation.
  \item We instantiate EWM with a \textbf{frontier language model} whose attributions are restricted to events that occurred in the trajectory.
\end{itemize}

\section{Related Work}
\label{sec:related}

\par\noindent\textbf{Credit assignment and reward redistribution.}
Standard RL propagates reward through bootstrapped or discounted returns~\citep{sutton2018rl,schulman2016gae}. Prior work bridges long temporal gaps using learned connections to relevant memories~\citep{ke2018sab,hung2019tvt}, return decomposition~\citep{arjonamedina2019rudder,patil2020aligned,zhang2023grd}, and outcome-conditioned or counterfactual attribution~\citep{harutyunyan2019hca,mesnard2021counterfactual,meulemans2023cocoa}. EWM shares the objective of assigning credit to relevant earlier decisions, but derives relevance from a prior-informed explanation of the realized trajectory rather than learning an attribution or return model exclusively from task interactions. It then conservatively redistributes the observed reward and applies an otherwise standard RL algorithm.

\par\noindent\textbf{Explanations and language-model rewards.}
Research in philosophy and cognitive science treats explanation as a mechanism for identifying the conditions responsible for an observed phenomenon and for supporting learning and causal inference~\citep{hempel1948explanation,gopnik1998explanation,keil2006explanation,lombrozo2006explanation,miller2019explanation}. Language models have also been used to generate reward functions and intrinsic goals~\citep{ma2024eureka,du2023ellm}. Most closely, LaRe uses an LLM to construct a latent-reward representation and learns a decoder that maps it to episodic returns~\citep{qu2025lare}. EWM instead asks the model which observed events explain a realized outcome, grounds its attributions in the trajectory, and reallocates an externally specified reward without defining a new reward function.

\par\noindent\textbf{World models and long-term recommendation.}
Conventional world models learn forward dynamics for planning or imagined rollouts~\citep{ha2018worldmodels,hafner2021dreamerv2,hafner2023dreamerv3}, while successor features separate reward-independent dynamics from task-specific value~\citep{barreto2017successor}. EWM is backward-looking and represents reward-independent explanatory relations among observed events. This attribution is complementary to work on long-term and generative recommendation, which primarily studies policy architecture, action representation, user modeling, and long-term value estimation~\citep{van2023estimating, deffayet2023generativeslate,gao2022kuairand}.

\section{Explanatory World Models}
\label{sec:method}
\begin{figure}
    \centering
    \includegraphics[width=0.9\linewidth]{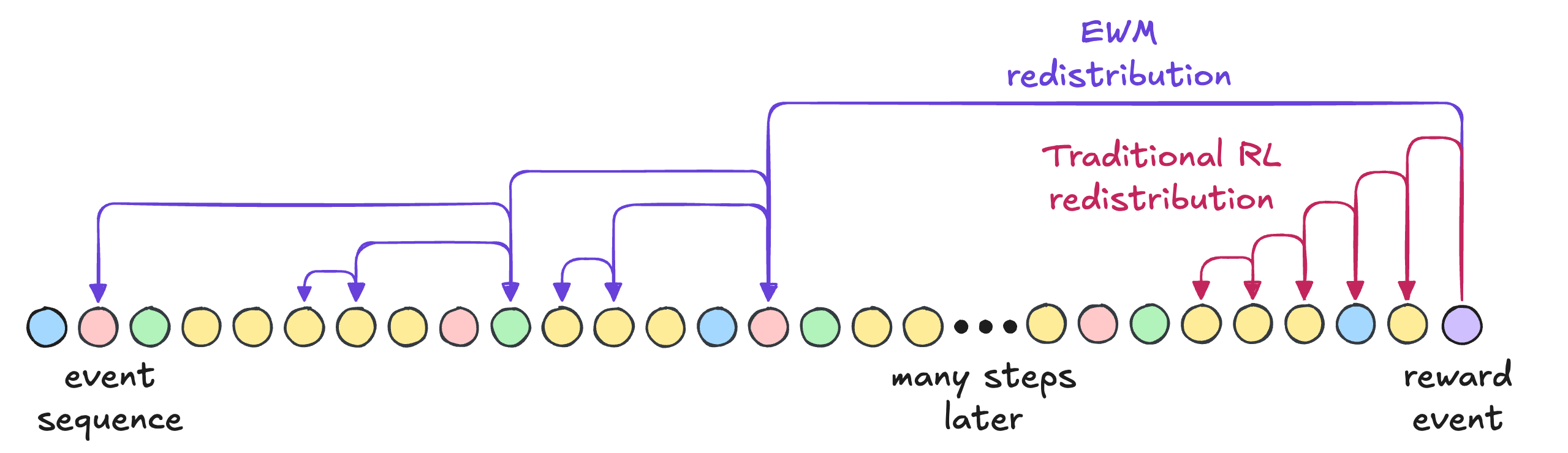}
    \caption{Traditional credit assignment (red) propagates a reward to temporally adjacent events, regardless of their relevance. EWM's conservative hierarchical operator (CHO, purple) instead uses the explanatory power of each event, recursively, to attribute the reward.}
    \label{fig:redist}
\end{figure}
This section presents Explanatory World Models and their use for credit assignment. We first define the general framework of explanations, then introduce the conservative hierarchical reward redistribution operator that translates explanations into learning signals, describe frontier models as a practical EWM instantiation, and finally integrate EWM to online and offline RL algorithms.

\subsection{What is an explanation?}
\label{sec:ewm}
Explanations take many forms, making a single definition difficult. \citet{miller2019explanation} describes explanation as an abductive process: given an observed event, or \emph{phenomenon}, an explainer infers a \emph{hypothesis} about the causes that brought it about. \citet{elster2007explaining} states this view succinctly: ``To explain a phenomenon is to cite an earlier event that caused it.'' Under this account, explanations express an agent's causal understanding of the world, and their adequacy depends on how faithfully that understanding captures the relations among events. EWM operationalize this idea by explaining observed outcomes and using the inferred relations to assign credit to preceding events.

As represented in Fig. \ref{fig:ewm}, an EWM takes a set of preceding events as input, which constitute the \emph{evidence}, and an observed event to be explained, which we call the \emph{phenomenon}. It returns a hypothesis that identifies the preceding events related to the phenomenon and assigns each one a weight representing its estimated contribution. We refer to this weighted hypothesis as an explanation.
The relations represented by an EWM are independent of the reward function: they describe how events in the environment relate, not how those events are valued. The same EWM can therefore be reused without learning those relations again.

We investigate how different operationalizations of explanation affect retrospective credit assignment. All variants receive the same observed history and retention outcome, but differ in the criteria used to select and assess explanatory events. We consider six variants, listed in Table \ref{tab:explanation-definitions}.
 All variants share the same input and output format but differ in their criteria for selecting and assessing explanatory events. HP is inspired by Halpern's account of actual causality ~\citep{halpern2005explanations}; D by Deutsch's hard-to-vary criterion \citep{deutsch2011beginning}; and LB and LS by Lombrozo's discussion of explanatory virtues ~\citep{lombrozo2006explanation, lombrozo2007simplicity}.
  
  \begin{table*}[!t]
  \centering
  \small
  \caption{Operationalization of explanations in each prompt variant.}
  \label{tab:explanation-definitions}
  \renewcommand{\arraystretch}{1.15}
  \begin{tabularx}{\textwidth}{@{}l p{0.20\textwidth} X@{}}
  \toprule
  \textbf{Variant} & \textbf{Criterion} & \textbf{Operationalization} \\
  \midrule

  B & Neutral explanation &
  Select plausible contributors to the outcome; return fewer or no events when evidence is weak. \\

  HP & Counterfactual relevance &
  Test whether removing or weakening an event makes the outcome less expected; omit redundant events on the same explanatory path. \\

  D & Hard-to-vary explanation &
  Require the explanation to depend on the specific history, behavior, and target. Replacing these details should weaken it; category matching alone is insufficient. \\

  LB & Explanatory breadth &
  Infer stable preferences or patterns supported across the history that explain and generalize beyond the outcome without covering unrelated behavior. \\

  LS & Explanatory simplicity &
  Prefer few underlying preferences or patterns; select representative events and distinguish foundational evidence from later reinforcement. \\

  LI & Integrated explanation &
  Combine breadth, simplicity, specificity, and behavioral grounding; explain the outcome, support generalization, and identify what would weaken the explanation. \\

  \bottomrule
  \end{tabularx}
\end{table*}
  
\subsection{Conservative hierarchical reward redistribution}
\label{sec:redist}

The \emph{credit-assignment problem} is the challenge of determining which actions, or sequences of actions, contributed to an observed outcome~\citep{minsky1961credit,sutton2018rl}. When an outcome produces a reward \(R\), the learner must translate that scalar signal into evidence about decisions made earlier in the trajectory. This becomes particularly difficult when rewards are sparse and the actions responsible for them occurred hundreds or thousands of steps earlier.
Reinforcement-learning algorithms commonly address temporal credit assignment using the discounted return
\begin{equation}
    G_t
    = \sum_{k=0}^{T-t-1}\gamma^k R_{t+k+1}
    = R_{t+1}+\gamma R_{t+2}+\gamma^2 R_{t+3}+\cdots,
\end{equation}
where \(\gamma\in[0,1]\) is the discount factor. A reward’s contribution to \(G_t\) decreases with its temporal distance from action \(a_t\). Discounting is computationally convenient, but it imposes a temporal prior: actions closer to a reward receive a stronger learning signal than actions farther away, regardless of their relevance to the outcome.
As illustrated in Figure~\ref{fig:redist}, temporally nearby events receive substantial initial credit even when they contributed little to the outcome, while the signal associated with distant but important decisions may become too weak to recover.

High-throughput simulators can partially compensate for this inefficiency by generating large numbers of trajectories. In real-world systems, however, interactions are expensive, delayed outcomes arrive slowly, and trajectories cannot be replayed freely. A production recommender may observe thousands of user events between a recommendation and its eventual effect on engagement or retention. EWM complements conventional RL by replacing temporal proximity as the sole initial attribution prior.

\paragraph{Conservative hierarchical operator (CHO)}
The EWM produces an explanation graph in which each phenomenon \(p\) is connected to a set of preceding events \(C_p=\{c_1,\ldots,c_K\}\), with attribution weights \(w_{p,c}\in[0,1]\). The redistribution operator uses this graph to determine where reward should appear in the trajectory. Unlike the fixed temporal coefficient \(\gamma^k\), these weights depend on the realized outcome and the evidence in the trajectory. Reward is therefore assigned according to explanatory relevance rather than temporal proximity. The policy optimizer may still apply a discount factor after redistribution, but it operates on the reward stream produced by EWM.

The redistribution is \emph{hierarchical} because an event cited in one explanation may itself be a phenomenon with an explanation. Starting from a rewarded phenomenon, the operator divides its reward among its eligible predecessors and recursively repeats this process for each predecessor that can be explained. Recursion terminates when it reaches a maximum depth \(D\), encounters an event without eligible predecessors, or the reward carried is below a defined threshold. The remaining reward is then deposited at that event.
At each node, we retain only predecessors whose weight exceeds a threshold \(\eta\), and normalize their weights locally:
\begin{equation}
    \bar{w}_{p,c}
    =
    \frac{w_{p,c}}
         {\sum_{c' \in C_p^{\eta}} w_{p,c'}},
    \qquad
    C_p^{\eta}
    =
    \{c \in C_p : w_{p,c}\geq\eta\}.
\end{equation}
If an amount \(m\) reaches \(p\), predecessor \(c\) receives the share \(m\bar{w}_{p,c}\). Consequently, the shares allocated at every node sum exactly to the amount received by that node.
The operator is \emph{conservative}: it changes the temporal location of reward but not its total amount.
\begin{algorithm}%
\caption{Reward Redistribution using Conservative hierarchical operator (CHO)}
\label{alg:ewm-redistribution}
\begin{algorithmic}[1]
\Require Reward sequence $r_{0:T-1}$; explanation graph $H$;
         maximum depth $D$; minimum attribution $\eta$
\State $\tilde{r} \gets r$
\ForAll{rewarded phenomena $p$ such that $r_p\neq 0$}
    \State $m \gets r_p$
    \State $\tilde{r}_p \gets \tilde{r}_p-m$
    \Comment{remove the mass before reallocating it}
    \State \Call{Allocate}{$p,m,D,\emptyset$}
\EndFor
\State \Return $\tilde{r}$
\Statex
\Function{Allocate}{$p,m,d,V$}
    \If{$d=0$ \textbf{or} $p\in V$}
        \State $\tilde{r}_p \gets \tilde{r}_p+m$
        \State \Return
    \EndIf
    \State $C \gets
        \{(c,w)\in H(p): c<p,\;c\notin V,\;w\geq\eta\}$
    \If{$C=\emptyset$ \textbf{or} $\sum_{(c,w)\in C}w=0$}
        \State $\tilde{r}_p \gets \tilde{r}_p+m$
        \State \Return
    \EndIf
    \State $W\gets\sum_{(c,w)\in C}w$
    \ForAll{$(c,w)\in C$}
        \State $m_c\gets m\,w/W$
        \State \Call{Allocate}{$c,m_c,d-1,V\cup\{p\}$}
    \EndFor
\EndFunction
\end{algorithmic}
\end{algorithm}

Note that by being recursive CHO can make reward reach arbitrarily far in the sequence of events even with partial trajectories. In other words, EWM and CHO do not need the full episode to reach far into the past, just intermediary events that link the reward event to the rest of the trajectory.

\subsection{Frontier models as EWM for RL}
\label{sec:ewm-rl}

\paragraph{Frontier models.}
We instantiate the EWM with a frontier language model, in our experiments \emph{Claude Sonnet 4.5}. Pretraining provides these models with broad priors about physical, procedural, and behavioral relations, while in-context reasoning allows them to apply that knowledge to previously unseen trajectories. We use the model as a backward-looking explainer rather than as a forward simulator, it observes both the completed trajectory and the realized phenomena, and identifies the preceding events that plausibly contributed to each phenomenon.

Conveniently, a single model invocation produces the direct explanation graph for the entire trajectory. Then, the produced graph is used by CHO to redistribute reward programmatically to cited events recursively up to the selected depth.
Recursion therefore requires no additional model calls, and the same cached explanation graph can be evaluated at different depths, or reward functions.

\paragraph{EWM for RL.}
EWM integrates with reinforcement learning as an attribution layer between observing rewards and updating the policy. Given a completed or partial trajectory $\tau$, we first obtain an explanation graph $\mathcal{H}_{\tau}$ and then apply the conservative hierarchical operator $\mathcal{R}$ to produce a redistributed reward sequence:
\begin{equation}
    \mathcal{H}_{\tau}
    = \operatorname{EWM}(\tau),
    \qquad
    \tilde{\mathbf{r}}
    = \mathcal{R}(\mathbf{r},\mathcal{H}_{\tau}).
\end{equation}
The RL algorithm subsequently computes returns and advantages from $\tilde{\mathbf{r}}$ rather than from the original reward sequence $\mathbf{r}$.
EWM therefore replaces a fixed notion of relevance based on temporal distance with trajectory-dependent attribution weights derived from the explanation graph.

In the online setting, the policy interacts with the original environment, once an episode is complete or enough steps are gathered, EWM explains its realized outcomes and redistributes the rewards before value targets and policy gradients are computed.
EWM does not otherwise modify the interaction protocol, policy architecture, or optimization algorithm, and can therefore be combined with other episodic RL methods that consume per-transition rewards.
For offline learning, EWM is applied once to each logged trajectory. The resulting explanation graphs and redistributed rewards can be cached and used to train a policy with return-weighted or advantage-weighted objectives.

\section{Long-Horizon RL Experiments}
\subsection{Crafter}
\label{sec:exp-crafter}
We first evaluate EWM in Crafter~\citep{hafner2022crafter}, a controlled environment with long, hierarchical dependencies between actions and achievements. This setting allows us to isolate the credit-assignment problem and test whether explanation-based redistribution improves learning as rewards become increasingly delayed. We compare EWM against PPO and LaRe~\citep{qu2025lare} baselines.

We use the symbolic JAX implementation of Crafter~\citep{matthews2024craftax} and aggregate the rewards of the technology achievements into a single terminal reward given at the end of the episode. By doing this we can assess how delayed and long-horizon rewards pose a challenge to RL methods.
We compare recurrent PPO and LaRe on this terminal reward stream with EWM at redistribution depth=1, with all methods using the same policy architecture and training configuration.

\begin{figure}[h]
    \centering
    \includegraphics[width=1.0\linewidth]{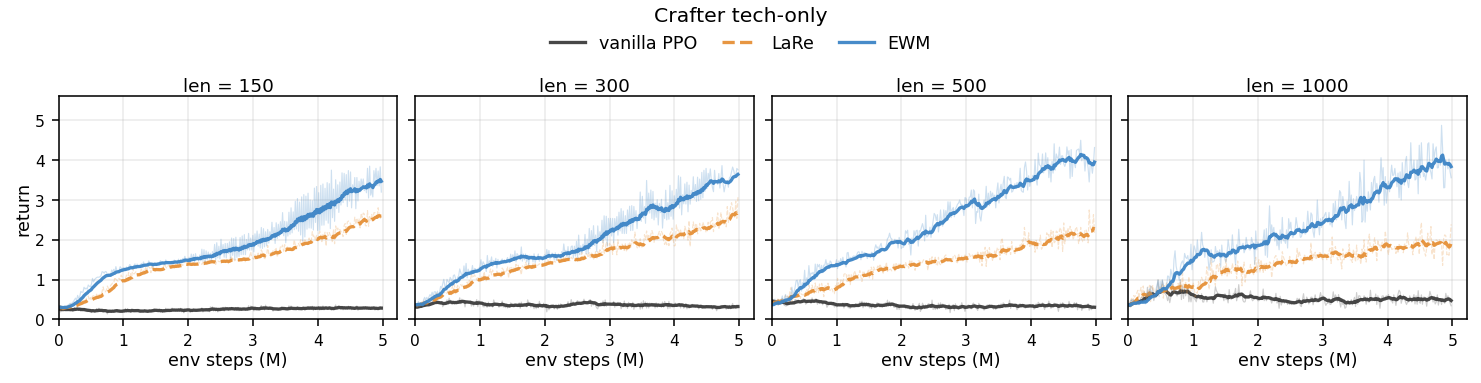}
    \caption{Crafter mean episode return at episode-length caps $L \in {150, 300, 500, 1000}$. Vanilla PPO stays near the random-policy floor. LaRe recovers some delayed credit; EWM is above LaRe at every horizon, and the gap widens as $L$ grows.}
    \label{fig:crafter}
\end{figure}

On Crafter, Figure~\ref{fig:crafter}, vanilla PPO clearly cannot learn in delayed long-horizon environments like this. The gap between EWM and LaRe widens with $L$ rather than with further training alone; at $L{=}150$ the two credit methods are still close, while at $L{=}1000$ EWM is about $2\times$ LaRe, consistent with the terminal reward becoming a longer-horizon assignment problem that redistribution is meant to solve. Moreover, in contrast to LaRe, EWM gets better with increases of $L$.

\subsection{WordCraft}
\label{sec:exp-wordcraft}
We next evaluate EWM on WordCraft~\citep{jiang2020wordcraft}, a combinatorial environment based on the Little Alchemy 2 game. Unlike Crafter’s largely fixed technology tree, WordCraft requires the agent to discover and compose recipes across a larger dependency graph, where early combinations may enable valuable discoveries several steps later.
We aggregate all rewards at the end of the episode, creating a controlled long-horizon setting in which the agent must assign a sparse terminal signal to the intermediate discoveries and actions that made it possible. WordCraft therefore tests whether EWM can recover compositional dependencies beyond the sequential progression studied in Crafter.

We train all methods using the same recurrent PPO architecture used for Crafter. At each mixing step, the agent selects two entities from a table of at most 24 items; a valid recipe produces a new entity that can support later combinations. We consider two difficulty levels that vary the number of distractors and the depth of the target recipe chain, with episodes capped at 50 mixing attempts. We evaluate a goal objective that provides a single reward when the target entity is created.
We compare vanilla PPO, LaRe, EWM-LLM, and additionally include EWM-GT, an oracle explainer that identifies the true recipe dependencies, as a diagnostic of the redistribution operator.

\begin{figure}[h]
    \centering
    \includegraphics[width=0.8\linewidth]{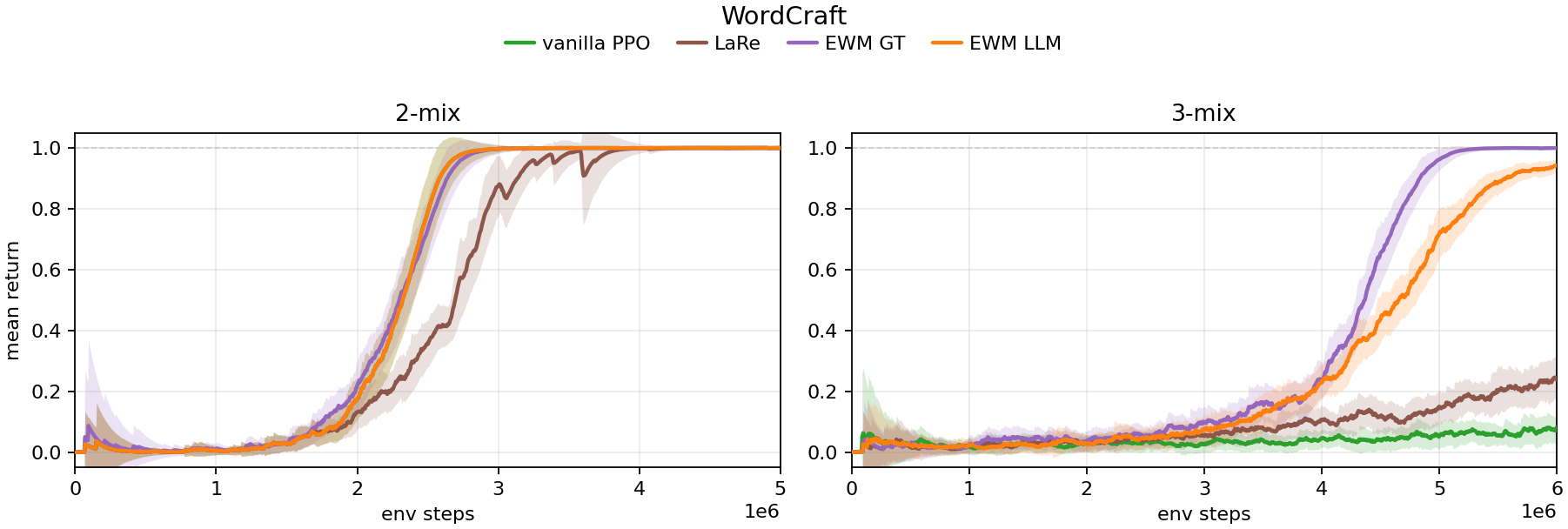}
    \caption{WordCraft returns. The 2-mix variant can easily be solved by every algorithm. The 3-mix variant is considerably more difficult due to longer causal chains and only EWM can solve it.}
    \label{fig:wordcraft}
\end{figure}

As shown in Fig. \ref{fig:wordcraft}, EWM provides its largest gains as the recipe chain becomes longer and the terminal reward becomes harder to assign. All models, EWM-LLM, PPO and LaRe reach the maximum return on the 2-mix split, whereas on the hard split EWM-LLM reaches almost oracle score, LaRe barely learns and PPO plateaus little above random.
These results indicate that the benefit of explanation-based redistribution emerges most clearly when success depends on assigning a sparse terminal reward across a deep sequence of compositional dependencies.

\section{Real-World Experiments}

Our empirical study covers the KuaiRand dataset in three item \footnote{In RL, the word ``episode'' normally stands for one full session of agent actions. In order to avoid any overload of language, we will name the agent recommendation as an ``item''.} recommendation settings. Firstly, we use EWM in a reward-weighted fine-tuning of Qwen2.5-1.5B-Instruct, where we compare the baseline LaRe method and six EWM explanation-variants from Sec. \ref{sec:method}. 
Then, we use the KuaiRand dataset in an online RL experiment, where we observe that EWM can accurately infer disengagement rules, thereby substantially improving cumulative return compared to vanilla methods like A2C~\citep{mnih2016asynchronous}.
Finally, we apply EWM to an industry-scale recommender .

\subsection{KuaiRand-offline}

In this section, we wish to answer the question: can explanation-based credit assignment improve recommendation learning from offline interaction data?       
Specifically, we study how different retrospective explanations affect reward-weighted supervised fine-tuning (SFT) on the KuaiRand dataset.

\vspace{-5mm}

\begin{figure}[h!]
    \centering
    \includegraphics[width=0.75\linewidth]{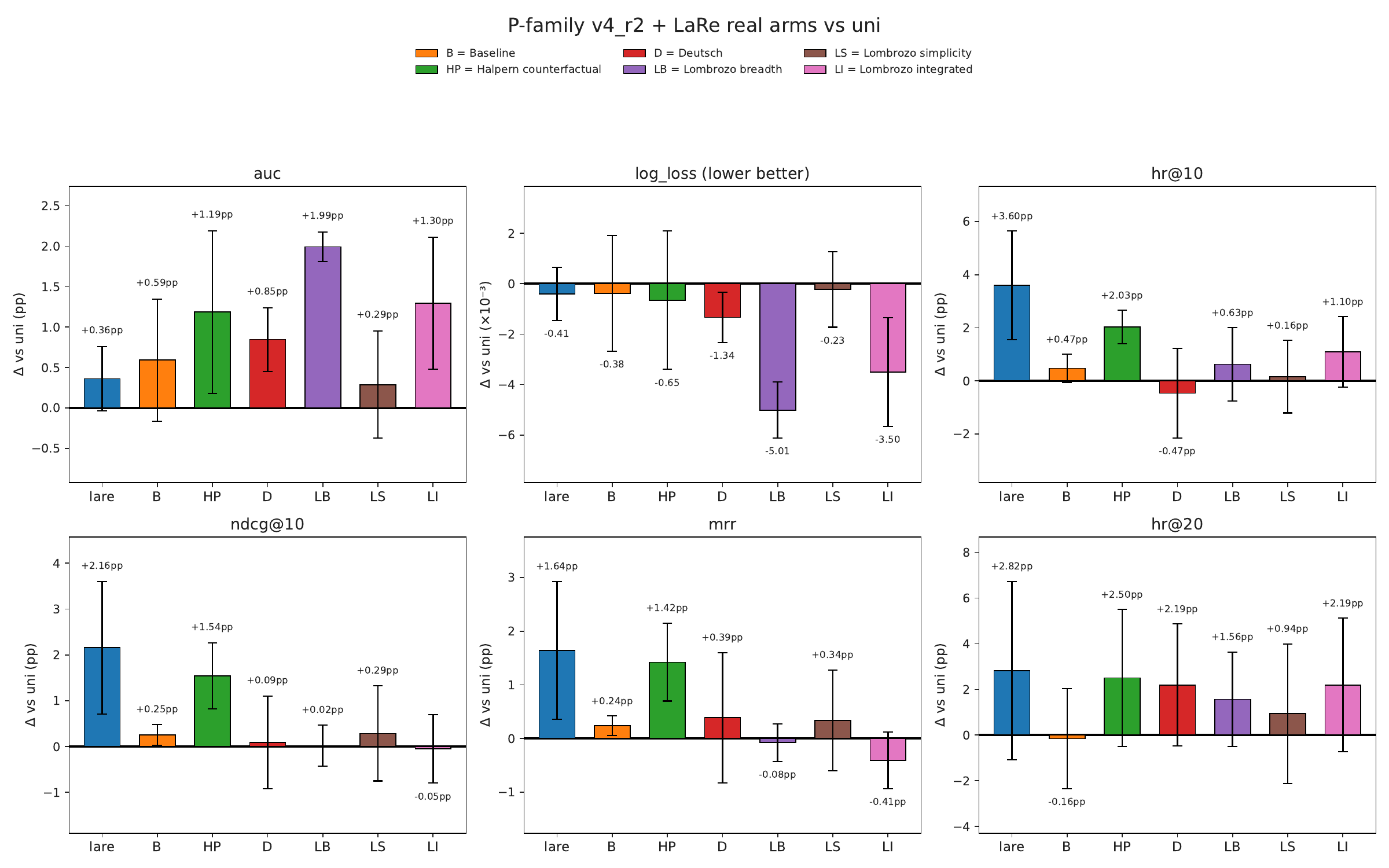}
    \caption{LB achieves the largest mean AUC improvement and log-loss reduction, whereas LaRe differs most from uniform on the four reported ranking metrics.}
    \label{fig:kuairandoffline}
\end{figure}

The results in Figure \ref{fig:kuairandoffline} provide evidence that explanation-based weighting can improve offline recommendation learning. All causal explanations yield positive mean AUC differences relative to Uniform. LB achieves the largest mean AUC improvement of 1.99 percentage points and the largest mean log-loss reduction of $5.01 \times 10^{-3}$. The ranking results reveal another pattern. LaRe achieves the largest mean differences across the four reported ranking metrics, while HP also changes all four metrics on average. In contrast, LB's mean NDCG@10 difference is close to zero and its mean MRR difference is slightly negative. This distinction is consistent with the different evaluation tasks: AUC measures discrimination across candidate pairs, whereas the ranking metrics evaluate the relative position among sampled candidates. Better aggregate discrimination and lower log loss, therefore, need not imply the trend for top-ranked recommendations; one method could substantially outperform the other on discrimination, yet still preserve the same policy when it comes to ranking recommendations. On the other hand, similar aggregate ranking metrics do not establish that two models produce similar candidate orderings. Intuitively, this suggests that LaRe and HP have resulted in significantly different ranking preferences compared to the uniform baseline, whereas LB's ranking preference stay relatively close to the baseline SFT.

\subsection{KuaiRand-online}

In this section, we wish to answer the question: can an explanation-derived behavioral rule improve sequential recommendation by preventing actions that cause user disengagement? We study this question in the KuaiRand environment using an A2C recommendation policy. The reward models, environment dynamics, policy architecture, and training configuration are shared across all experiments. In this experiment, the environment has a hidden termination rule that terminates an episode depending on tag overlap between a candidate video and previously recommended videos, simulating a bored user. We compare four conditions: the vanilla A2C policy, an EWM-derived five-step termination rule, a full-history extension of the EWM-derived rule, and an oracle policy whose recommendations are constructed from the environment's actual termination rule.

\begin{figure}[h!]
    \centering
    \includegraphics[width=\linewidth]{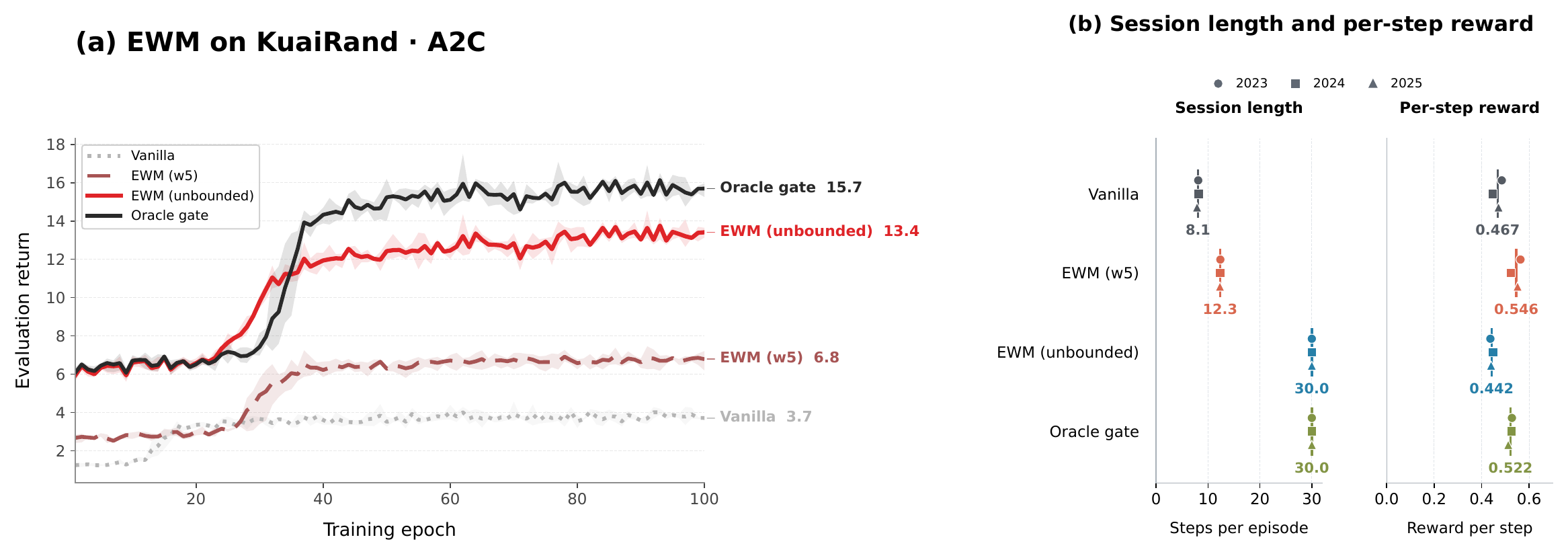}
    \caption{The termination rule that 5-window-EWM discovered outperforms the baseline vanilla A2C method, and the full-history EWM comes close to the oracle termination rule results.}
    \label{fig:kuairandterm}
\end{figure}

Explanation-derived action gating improves cumulative return in this simulated recommendation setting. Mean evaluation return increases from 3.773 for Vanilla A2C to 6.736 for EWM (w5) and 13.249 for EWM (unbounded), compared with 15.653 for the Oracle gate. These results show that an approximate explanation-derived constraint can be useful even without exactly recovering the environment's termination mechanism.

The session-length and per-step-reward measurements clarify
the sources of these gains. EWM (w5) increases mean session
length from approximately 8.1 to 12.3 recommendations while
also increasing mean reward per step from 0.467 to 0.546.
The full-history extension reaches the 30-step horizon,
but its mean reward per step decreases to 0.442. Its larger
cumulative return is therefore accompanied by longer
interaction rather than higher average reward per
recommendation.

\subsection{Industry Generative Recommender}
\label{sec:exp-lpm}

We evaluate EWM on an industry-scale generative recommender from a global streaming service, where recommendations may influence engagement days later. We test whether \emph{explanatory reach} can capture this long-term influence without prescribing a specific engagement metric.
For each logged recommendation, a frontier language model constructs a recommendation-rooted explanatory graph over the user’s realized interactions. 
We derive a reward from this graph, \emph{Reach}, defined as the number of distinct downstream events connected to the recommendation through high-confidence explanatory relations.

We evaluate by offline replay against a held-out log of past recommendations. Each policy produces a slate for every logged context, and we record whether it reproduces the recommendation that was actually served, together with that recommendation's realised outcome and scoring a miss as zero. We report paired contrasts against the SFT baseline with bootstrap intervals.

\begin{table}[!h]
\centering
\small
\caption{Offline evaluation of EWM-reach RLFT against the SFT baseline. \emph{Recall} is the rate of reproducing the logged recommendation; \emph{converted} and \emph{broad reach} additionally require that action to be followed by a further stream or two. All rows are fold-changes against the baseline.}
\setlength{\tabcolsep}{5pt}
\begin{tabular}{@{}lc@{}}
\toprule
Metric & EWM RLFT vs baseline ($95\%$ CI) \\
\midrule
Recall & $3.75\times\ [2.62, 5.79]$ \\
Converted & $3.86\times\ [2.11, 10.0]$ \\
Broad reach ($\geq$2 streams) & $4.00\times\ [2.18, 11.5]$ \\
\bottomrule
\end{tabular}
\label{tab:lpm-reach}
\end{table}

\section{Conclusion}
\label{sec:conclusion}

We introduced Explanatory World Models (EWM), which address long-horizon credit assignment by explaining realized outcomes and redistributing reward to relevant earlier events.
Our conservative hierarchical operator redistributes credit to relevant events identified by the EWM.
We instantiated EWM with a frontier language model and integrated it with both online and offline learning.
Across Crafter and WordCraft, EWM outperforms PPO and LaRe under sparse rewards. It also improves over SFT and LaRe in offline KuaiRand experiments and over PPO online, while increasing recall and downstream engagement measures in an industry-scale generative recommender.
These results show that explanations can provide structured credit when temporal propagation becomes inefficient.

\bibliography{refs}
\bibliographystyle{iclr2027_conference}

\clearpage
\appendix
\section{EWM Redistribution Analysis}
\label{app:redist}

\begin{figure}[!h]
    \centering
    \includegraphics[width=\linewidth]{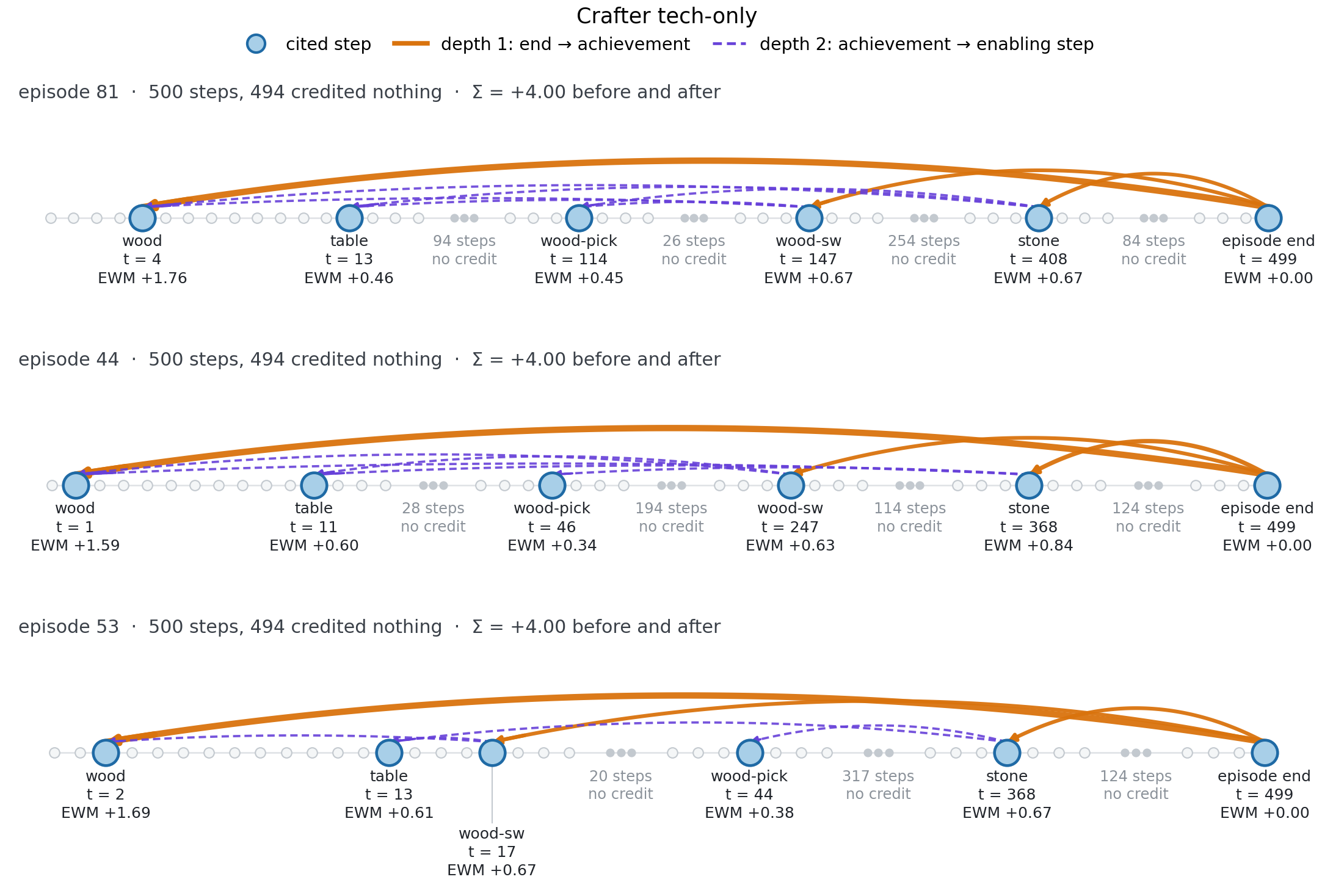}
    \caption{Three Crafter episodes with the same conserved return. Each row is that episode's step axis, compressed: cited steps keep a few real neighbours, and each long uncredited run collapses to three dots labelled with the number of hidden steps. Solid orange arrows are depth-$1$ hops from the episode end onto the achievements that earned the reward; dashed purple arrows are depth-$2$ hops from those achievements onto earlier enabling actions.}
    \label{fig:crafter-redist}
\end{figure}

Figure~\ref{fig:crafter-redist} is a static view of Algorithm~\ref{alg:ewm-redistribution} on three stone-reaching episodes from a dump (terminal reward, $L{=}500$, $d{=}2$). It is not a training curve: the explainer is Claude Sonnet 4.5 and the arrows are the cited causes.

The episode end is a phenomenon holding the whole return ($+0.5$ wood $+1.5$ wood-sword $+2.0$ stone $= +4.00$). Depth $1$ (orange) moves that mass onto the three achievements the explainer names as causes of the lump. Depth $2$ (purple) then asks, for each of those achievements, what enabled \emph{it}.
Those second hops are the only reason \texttt{place table} and \texttt{make wood pickaxe} receive credit. A depth-$1$ operator would have left them at zero. Wood is cited both by the lump and by later achievements, which is why it receives the largest share (about $+1.6$ to $+1.8$).
Under each episode end the environment still records $+4.00$ and EWM has moved the mass, not added any. Summing the EWM amounts on the blue circles recovers $+4.00$ in every row.

\section{Experiment Details}
\label{app:exp-details}

\subsection{KuaiRand-offline}

We compare LaRe and six EWM-prompt-variants against a uniform-weight baseline. All experiments use the same Qwen2.5-1.5B-Instruct~\citep{yang2024qwen25} model initialization, training recommendation samples, and training procedure. The experimental intervention is the weighting of supervised training examples derived from the assigned credit.

\subsubsection{Dataset and preprocessing}

We use a preprocessed KuaiRand dataset containing chronological user-video interaction sequences. Binary engagement labels on the feed histories indicate whether an interaction includes a like, follow, forward, or long-view signal.
The experiment uses a fixed subset of 1,000 user sessions from the preprocessed data. Following the stored session order, the first 700 sessions are used for training and the remaining 300 for evaluation. This is a fixed session-level partition, rather than a new random split for each training seed. Each evaluation session includes a held-out target from the randomly exposed interactions.
Videos are represented through textual metadata, including available captions, creator information, and category descriptions. The implementation reuses a shared recommendation-data interface adapted for KuaiRand videos.

\subsubsection{Recommendation model and input representation}

All experiments fine-tune the Qwen2.5-1.5B-Instruct model. Each supervised example contains the user's preceding interaction history and a candidate video. The target response is either \texttt{<answer>like</answer>} or \texttt{<answer>skip</answer>}, according to the recorded binary engagement label.
The credit-assignment method changes the weight of an example, while its input and target label remain unchanged. In other words, an explanation does not replace the observed engagement label with an LLM-generated label.
The maximum input sequence length is 4,096 tokens. Training uses the model's chat template, and the supervised loss is applied to answer tokens rather than to the history and candidate-description tokens.

\subsubsection{Credit-assignment mechanisms}
We compare eight methods:
\begin{itemize}
  \item Uniform, which assigns unweighted SFT.
  \item LaRe, which uses the LaRe credit annotations to construct training weights.
  \item B, a baseline EWM prompt.
  \item HP, a prompt emphasizing Halpern-style counterfactual explanation.
  \item D, a prompt emphasizing Deutsch-style explanation.
  \item LB, a prompt emphasizing Lombrozo-style explanatory breadth.
  \item LS, a prompt emphasizing Lombrozo-style explanatory simplicity.
  \item LI, an integrated EWM prompt combining LB and LS.
\end{itemize}

For the explanation-prompt variants, the annotation task concerns user retention. The annotator receives a chronological video history and an observed retention outcome, expressed through the user's subsequent return activity. It identifies videos in the history that plausibly contributed to that outcome.
The prompt requests strict JSON containing selected step indices, importance scores, and short explanations. At most five contributing steps are requested. Importance scores lie between zero and one and are not required to sum to one.
Let $R_e$ denote the nonnegative retention outcome for session $e$, and let $a_{e,t}$ denote the valid positive importance assigned to step $t$. The corresponding per-step reward is
\begin{equation}
  \widetilde r_{e,t}
  =
  R_e
  \frac{a_{e,t}}{\sum_j a_{e,j}},
\end{equation}
when at least one valid positive importance is available. Unselected steps receive zero reward. If no valid contributing step is available, the reward remains at the final step. Thus, this redistribution preserves the undiscounted session outcome:
\begin{equation}
  \sum_t \widetilde r_{e,t}=R_e.
\end{equation}

For each annotated session, we compute discounted returns from the redistributed rewards:
\begin{equation}
  G_{e,t}
  =
  \sum_{k=t}^{T_e-1}
  \gamma^{k-t}\widetilde r_{e,k},
  \qquad \gamma=0.95.
\end{equation}
Returns are converted into positive training weights using an exponential transformation:
\begin{equation}
  q_{e,t}
  =
  \exp\!\left(
      \operatorname{clip}
      \left(
          \frac{G_{e,t}}{\tau},-20,20
      \right)
  \right),
  \qquad \tau=5.
\end{equation}
The weights are then normalized jointly across the training examples:
\begin{equation}
  w_{e,t}
  =
  \frac{q_{e,t}}
  {\frac{1}{N}\sum_{e',t'}q_{e',t'}+\epsilon},
  \qquad \epsilon=10^{-8}.
\end{equation}
The model minimizes weighted answer-token cross-entropy:
\begin{equation}
  \mathcal{L}(\theta)
  =
  -\frac{1}{|\mathcal{B}|}
  \sum_{(e,t)\in\mathcal{B}}
  w_{e,t}
  \sum_{\ell\in\mathcal{A}_{e,t}}
  \log p_\theta
  \left(
      y_{e,t,\ell}
      \mid x_{e,t},y_{e,t,<\ell}
  \right),
\end{equation}
where $\mathcal{A}_{e,t}$ indexes the answer tokens. For Uniform, $w_{e,t}=1$ for every example.
This is an offline supervised objective. It does not use PPO updates, an online recommendation simulator, or a learned value-function baseline.

\subsubsection{Training budget}

We use seeds 1, 2, and 3. Each condition is trained for one epoch with batch size 4 and AdamW at an initial learning rate of $10^{-5}$. Weight decay is zero, gradients are clipped to a norm of 1.0, and training uses bfloat16 precision.
The learning-rate schedule uses a warm-up fraction of 0.03, followed by cosine decay to 10\% of the initial learning rate. The saved final evaluation snapshots for all conditions occur at optimization step 5,423.

\subsubsection{Evaluation protocol}

Evaluation measures held-out engagement prediction and candidate ranking. For each history--candidate pair, the model produces an engagement score from a two-way softmax over the next-token logits
corresponding to \texttt{like} and \texttt{skip}:
\begin{equation}
  s_\theta(h,i)
  =
  \frac{\exp(z_{\mathrm{like}})}
  {\exp(z_{\mathrm{like}})+\exp(z_{\mathrm{skip}})}.
\end{equation}
We report AUC and binary log loss over 300 held-out user targets. Ranking evaluation is restricted to the 213 users whose held-out target has a positive engagement label. For each eligible user, the positive target is ranked against 99 sampled unseen videos using the same model score. We report HR@10, HR@20, NDCG@10, and mean reciprocal rank.
The figure uses the final evaluation snapshot. For each metric, we first compute the within-seed difference between a method and Uniform:
\begin{equation}
  \Delta_{m,s}
  =
  M_{m,s}-M_{\mathrm{Uniform},s}.
\end{equation}
Bars show the mean of these differences across the three seeds. Error bars show one standard error, computed as the sample standard deviation of the three paired differences divided by $\sqrt{3}$.
AUC and ranking-score differences are expressed in percentage points. Log-loss differences are expressed in units of $10^{-3}$.

\subsection{KuaiRand-online}

\subsubsection{Dataset and sequential recommendation environment}

We use the KuaiRand environment provided by EasyRL4Rec~\citep{yu2024easyrl4rec}, with \texttt{is\_click} as the feedback target. Policy training uses the DeepFM user model to generate simulated feedback. Evaluation uses the MF-based reward matrix.
We follow the reward-model architecture and principal training configurations provided by EasyRL4Rec~\citep{yu2024easyrl4rec}, using a five-member DeepFM ensemble for simulated training feedback and an MF-based rating matrix for policy evaluation~\citep{easyrl4rec_movielens_example, easyrl4rec_movielens_mf_results, easyrl4rec_usermodel_utils}.
These scores are model-based surrogates for user feedback. An evaluation return therefore measures performance in the simulated recommendation environment rather than directly observed engagement from an online deployment.
At the beginning of an episode, the environment samples a user and initializes the recommendation state. The policy then repeatedly selects a video, receives feedback, and updates its state. Episodes end when the environment's disengagement condition is met or when the maximum horizon of 30 recommendations is reached.
The disengagement mechanism depends on repeated item tags within a history window. The configured parameter \texttt{num\_leave\_compute} is 10 and the leave threshold is zero. The oracle gate reproduces the implementation's exact history-window construction. The environment and its termination mechanism remain the same across all four conditions.

\subsubsection{A2C policy}

The recommendation policy uses a shared two-layer network with hidden sizes 64 and 64, followed by an actor head and a value head. The actor produces a categorical distribution over candidate videos, and the critic estimates the return from the current state.
The actor--critic and state tracker are optimized with separate Adam optimizers, both using learning rate 0.001. The discount factor is 0.9, the generalized advantage estimation parameter is 1.0, the value-loss coefficient is 0.5, and the entropy coefficient is zero. No explicit maximum gradient norm is configured in these runs.
Training actions are sampled from the categorical policy. Evaluation uses the implementation's deterministic action-selection mode.

\subsubsection{Action-gating conditions}

Let $\mathcal{T}(i)$ denote the tags associated with candidate item $i$, and let $\mathcal{H}_t^{(m)}$ denote the history selected by gating condition $m$. A recommendation candidate is normally permitted when its tags do not overlap with the tags in that history:
\begin{equation}
  M_t^{(m)}(i)
  =
  \mathbb{I}\left[
      \mathcal{T}(i)
      \cap
      \bigcup_{j\in\mathcal{H}_t^{(m)}}\mathcal{T}(j)
      =
      \varnothing
  \right].
\end{equation}
The resulting action distribution is
\begin{equation}
  \pi_\theta^{(m)}(i\mid s_t)
  =
  \frac{
      M_t^{(m)}(i)\pi_\theta(i\mid s_t)
  }{
      \sum_j M_t^{(m)}(j)\pi_\theta(j\mid s_t)
  }.
\end{equation}
The four conditions are:
\begin{itemize}
  \item Vanilla A2C policy.
  \item EWM (w5), which discovers a rule that blocks candidates sharing a tag with any of the five most recent recommendations. This method implements the discovered rule.
  \item EWM (unbounded), which extends the same tag-overlap rule to the entire recommendation history of the current episode.
  \item Oracle gate, which uses the environment's actual disengagement-rule window.
\end{itemize}
The discovered rule is fixed before the reported A2C training runs. It is executed as an action mask, so no LLM call is required for each recommendation.
The gate is applied during both training rollout collection and evaluation. The corresponding mask is also used when computing action probabilities during the A2C update.
The gate does not alter the numerical reward assigned to a selected action. However, because it changes the selected actions and resulting trajectories, realized episode returns need not be equal across conditions.

\subsubsection{Training budget}

We use seeds 2023, 2024, and 2025. Each condition is trained for 100 epochs with a target of 10,000 environment steps per epoch. Training uses 100 parallel environments, collects 100 episodes per collection round, and uses minibatches of size 1,024 with one optimization repeat per collection round. All four conditions have complete 100-epoch records for all three seeds.

\subsubsection{Evaluation protocol}

Our primary results use the standard free-running evaluation protocol, in which episodes terminate according to the environment's disengagement rule or the 30-step horizon. Each recorded epoch-level evaluation contains 100 episodes. Alternative evaluation variants recorded by the framework are not pooled into the reported results.
For each seed, we summarize performance using the mean evaluation return over epochs 76--100:
\begin{equation}
  \overline{R}_s
  =
  \frac{1}{25}
  \sum_{e=76}^{100} R_{s,e}.
\end{equation}
We report the mean and sample standard deviation of these seed-level summaries.
To examine the source of cumulative-return differences, we additionally report mean episode length and reward per step over the same window. The latter is computed as
\begin{equation}
  Q_s
  =
  \frac{
      \sum_{e=76}^{100}R_{s,e}
  }{
      \sum_{e=76}^{100}L_{s,e}
  },
\end{equation}
where $L_{s,e}$ is the mean episode length. Since every evaluation contains the same number of episodes, this ratio equals total evaluation reward divided by total recommendation steps within the window.
Learning curves show unsmoothed mean returns across the three seeds. Shaded bands indicate the minimum--maximum seed range, not confidence intervals.

\subsubsection{Results}

The mean evaluation returns are $3.773 \pm 0.143$ for Vanilla, $6.736 \pm 0.269$ for EWM (w5), $13.249 \pm 0.184$ for EWM (unbounded), and $15.653 \pm 0.252$ for the Oracle gate, where the uncertainty denotes the sample standard deviation across the three seed-level summaries. The ordering is consistent across all three seeds.
Relative to Vanilla, EWM (w5), EWM (unbounded), and the Oracle gate achieve approximately $1.79\times$, $3.51\times$, and $4.15\times$ the cumulative return, respectively.

The episode-length and per-step-reward measurements clarify these differences. Vanilla averages approximately 8.1 recommendations per episode, compared with 12.3 for EWM (w5). Both EWM (unbounded) and the Oracle gate reach the 30-step horizon throughout the final evaluation window. The discrepancy between EWM (unbounded) and the Oracle gate is likely because EWM (unbounded) prohibits recommendation items that are actually allowed, hence missing out on certain high-reward recommendations.

The corresponding mean seed-level rewards per step are approximately 0.467, 0.546, 0.442, and 0.522. Thus, the unbounded gate's larger cumulative return is accompanied by substantially longer sessions, despite lower average reward per step than Vanilla. This is because the vanilla policy only optimizes for greedy recommendation rewards. The Oracle gate achieves the same session length as the unbounded gate but obtains higher reward per step.
These results support the usefulness of explanation-derived rules as operational constraints in this simulated recommendation setting. They also distinguish two aspects of recommendation performance:
sustaining interaction and obtaining higher reward from each recommendation. The gains apply to the gated policy system and should not be interpreted as evidence that reward redistribution alone
produces the same effect.

\subsection{MovieLens-1M}

In this section, we wish to answer the question: will EWM help in an online recommendation setting? Specifically, we study temporal credit assignment in a sequential recommendation setting using the MovieLens-1M dataset. The policy architecture, environment dynamics, and total reward are held fixed across all experimental conditions. The only experimental variable is how the reward of a completed recommendation session is distributed across individual recommendation actions before computing PPO returns and advantages. We compare four algorithms: dense ground truth rewards, terminal reward only, uniform redistribution, and EWM-based credit redistribution, which is an informed retrospective allocation of episode-level outcomes.

\subsubsection{Dataset and preprocessing}

We use the MovieLens-1M dataset, which contains one million explicit ratings from six thousand users across three thousand movies. The ratings lie on a five-point scale. We split the dataset using a global 90/10 split with random seed 42. This is neither a chronological split nor a per-user split, and the training set has nearly the same coverage as the validation set.

User features consist of user ID, gender, age group, and occupation. Movie features consist of movie ID and up to six genre fields. All categorical fields use eight-dimensional embeddings. The resulting concatenated user representation has 32 dimensions, while the concatenated movie representation has 56 dimensions. The environment uses dense zero-based user indices and movie indices.

\subsubsection{Training and evaluation reward models}

The experiment uses two different dense reward matrices:
\begin{itemize}
    \item Five-model DeepFM ensemble, which is used to simulate online feedback
    \item Six-factor matrix factorization, which is used only for policy evaluation and not for training, i.e. an independent surrogate evaluator
\end{itemize}

\paragraph{DeepFM training simulator} The training simulator is a five-member DeepFM ensemble. DeepFM predicts a rating from a static user–movie feature tuple and does not read the user’s sequential viewing history. For a given feature vector, its predicted mean combines a linear term, factorization-machine interactions, and a nonlinear network:

\begin{equation}
    \widehat R_{\mathrm{DF}}(u,i)
    b+w^\top x_{ui}
    +\sum_{p<q}\langle v_p,v_q\rangle x_{ui,p}x_{ui,q}
    +f_\theta(E(x_{ui})).
\end{equation}

The network contains two 128-unit ReLU layers. The ensemble is trained for five epochs using Adam, batch size 2048, embedding dimension 8, and a L2 regularization coefficient of 0.1. After training, each member predicts all possible user–action combinations. The five predicted means are averaged to obtain a dense user x action matrix.

\paragraph{MF-based evaluation simulator} Policy evaluation uses a separate matrix-factorization model with six-dimensional user and movie factors. It is trained on the same 90\% interaction split using mean squared error, SGD with learning rate 0.1 and momentum 0.9, for 80 epochs. Its predictions for all user-action pairs are clipped between $0$ and $5$ before being used by the environment. Consequently, an RL test return is a cumulative score from an MF-completed surrogate environment.

\paragraph{What is the difference between DeepFM and MF?} DeepFM and matrix factorization (MF) are fitted on the same 90\% training split, but differ in their inputs, architectures, and optimization procedures. DeepFM uses user ID, gender, age group, and occupation, together with movie ID and genre features. Each categorical field has an eight-dimensional embedding. The model combines a linear component, factorization-machine interactions, and a nonlinear network with two 128-unit ReLU layers. Five DeepFM models are trained for five epochs each using Adam and batch size 2,048; their predictions are averaged to construct the dense training reward matrix.
In contrast, MF uses only user and movie identities. It learns six-dimensional user and movie factors and predicts ratings through their inner products. The model is trained for 80 epochs using mean squared error and SGD with learning rate 0.1 and momentum 0.9. Its predictions form the dense evaluation reward matrix.
Before use by the environment, the DeepFM ensemble predictions are shifted by subtracting their global minimum, whereas MF predictions are clipped to $[0,5]$. Optimizing dense DeepFM rewards does not necessarily maximize MF evaluation return. Since both models use the same interaction split, MF provides an alternative model-based evaluator rather than an independent source of ground-truth user feedback.

\subsubsection{Sequential recommendation environment}

At the beginning of each episode, a user is sampled uniformly from all 6,040 users. The environment observation always consists of the user index and the most recent recommendation item. With random initialization enabled, the first observation contains a randomly initialized movie and zero feedback; this initial movie is not an action selected by the policy. The policy then acts by selecting one movie to recommend. Episodes have a maximum horizon of 30 actions.

\subsubsection{Actor input and state representation}

Thus, the actor receives a 57-dimensional state containing the average movie representation and average recent feedback. The pretrained movie embeddings are initialized from the first DeepFM ensemble member. At the first action, the state consists of the randomly initialized movie embedding concatenated with zero feedback.

\subsubsection{PPO policy}

All algorithms use the same discrete PPO actor–critic network. The network has a shared two-layer MLP that maps the 57-dimensional state through hidden layers of sizes 64 and 64 with ReLU activation. Two separate heads of the network produce: (1) a masked categorical distribution over all actions; (2) a scalar estimate for the possible future returns expected from the current state. Training actions are sampled from the categorical policy, whereas evaluation uses the deterministic argmax action. Stochasticity during training comes from categorical sampling. Advantages are normalized within each minibatch. Gradients are clipped to a norm of 0.2. Both the actor–critic and state tracker are optimized with Adam at learning rate 0.001.

\subsubsection{Temporal credit-assignment conditions}

For a completed trajectory, we construct the PPO learning reward in four ways:

\begin{itemize}
    \item Dense, where the reward for each movie recommendation is passed to the PPO algorithm
    \item Lump, where the algorithm only observes the terminal reward, i.e. the sum of all dense rewards
    \item Uniform, where the terminal reward is distributed uniformly across all past recommendation actions
    \item EWM, where the reward assigned to each step is inferred by the EWM framework
\end{itemize}

\subsubsection{EWM credit redistribution}

In the current instantiation, EWM uses claude-sonnet-4-5 as a retrospective credit allocator. For each completed episode, the prompt contains three types of information:

\begin{itemize}
    \item First, a user profile is constructed exclusively from the training split, containing information such as gender, age group, occupation, etc.
    \item Second, each recommended movie (i.e. past actions) is described using its title, genres, training-set average rating, and number of training raters. This is so that the EWM can have context on what the recommendation task looks like.
    \item Third, EWM receives a scalar episode outcome
\end{itemize}

Crucially, the individual step rewards are never included in the prompt to the EWM. The EWM sees only the aggregate terminal reward. The EWM must return strict JSON, assigning the terminal reward to every recommended episode. Invalid or repeatedly unsuccessful judgments do not zero the trajectory; instead, the original Lump allocation is retained.

Successful judgments are stored in an append-only cache keyed by the prompt version, user, and exact ordered movie sequence. RL training runs share this cache and are run sequentially. This improves efficiency and ensures that an identical cached trajectory receives the same credit allocation, and also reduces the noise from LLM judgment. The current configuration uses eight concurrent requests, a limit of 100 attempts per minute, a 240-second timeout, up to 2,400 output tokens, and a budget of 45,000 fresh episodes per seed. Two API keys alternate requests to the same model and endpoint; they are a throughput mechanism, not two independent judges. Temperature is not explicitly specified.

\subsubsection{Training budget}

We use seeds 2023, 2024, and 2025. Each condition is trained for 100 epochs and 100 episodes per epoch. PPO uses batch size 1,024 and one optimization pass per minibatch.

\subsubsection{Evaluation protocol}

Evaluation uses the MF-based environment and deterministic action selection. One evaluation is performed before training and one after every epoch. Each evaluation contains 100 episodes. Additional metrics include episode length, mean reward per turn, catalog coverage, intra-session diversity, and novelty. For the main table, we recommend reporting the epoch-100 return for each seed, followed by the mean and sample standard deviation across the three seed-level means.

\subsubsection{Results}

\begin{figure}[h!]
    \centering
    \includegraphics[width=0.7\linewidth]{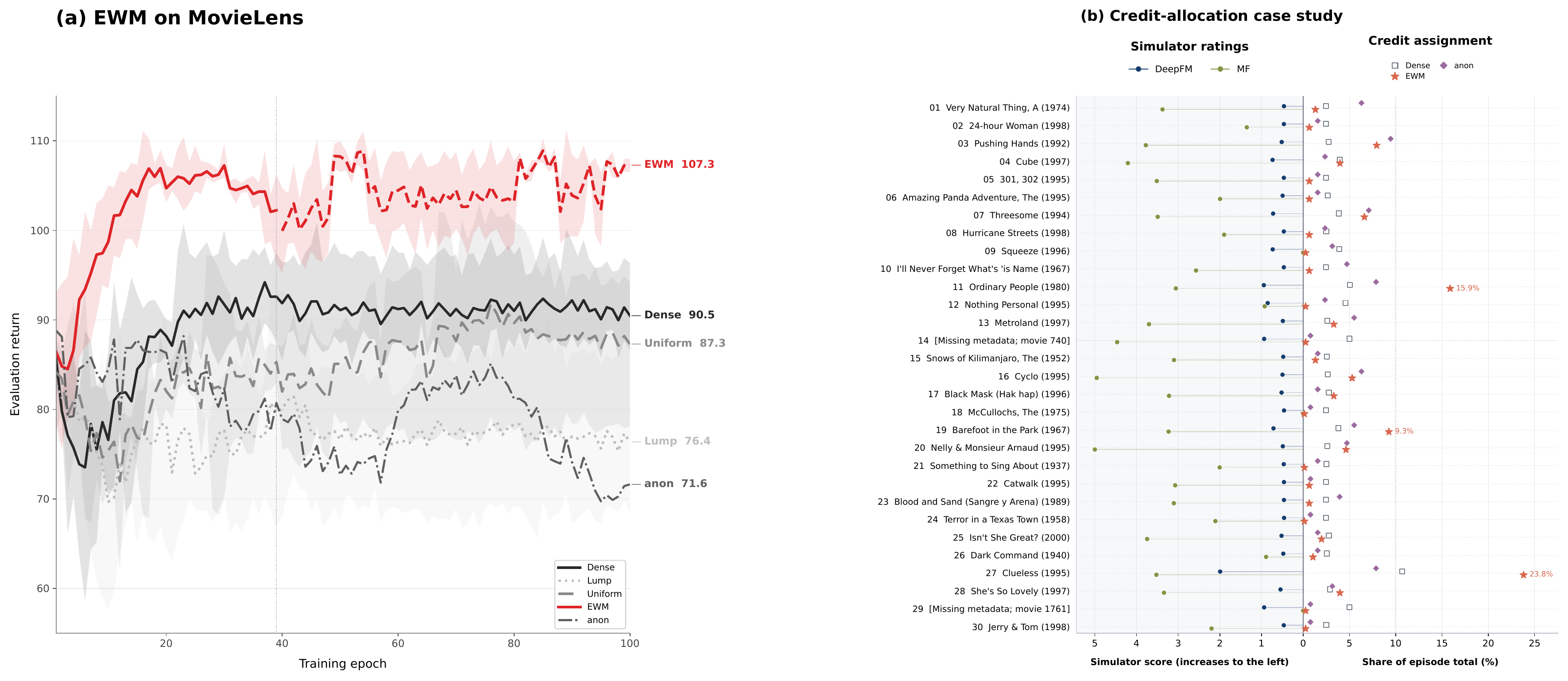}
    \caption{MovieLens-1M policy learning and an illustrative credit allocation. (a) Mean evaluation return across all seeds; shading shows the seed minimum–maximum range, not confidence intervals. We see that the performance of EWM outperforms baselines dense feedback, uniform redistribution, lump terminal reward, and an ablation where we anonymize all the item names in the EWM prompt (so that the EWM does not get any information about the user). Here, each standard deviation is across 100 evaluation episodes. (b) An example session (user 4707, row 573; 30 recommendations in original order), where training rewards and evaluation scores increase to the left. We can see that EWM predictions happen to line up quite well with MF predictions, especially when MF and DeepFM issues different predictions about the same episode.}
    \label{fig:movielens}
\end{figure}

\begin{figure}[h!]
    \centering
    \includegraphics[width=0.7\linewidth]{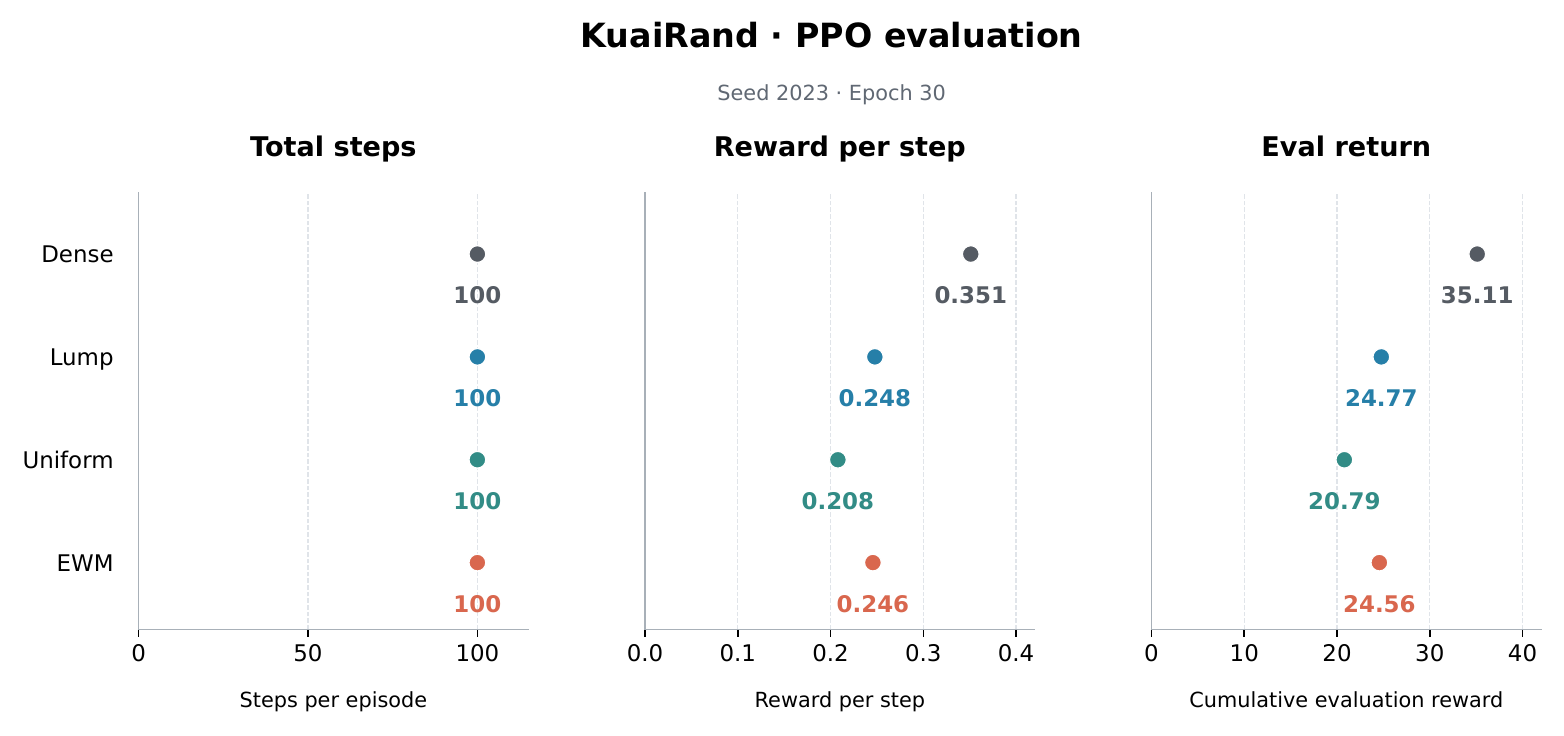}
    \caption{Anonymized KuaiRand PPO results for Dense, Lump, Uniform and EWM. Each point is a statistic from the same 100-episode evaluation. Total steps denotes the mean steps per episode, not training steps or the sum across evaluation episodes. Eval return is mean cumulative reward per episode. Similar to the anonymization ablation experiment in the MovieLens setting, we again see here that EWM performs worse than dense baselines in the KuaiRand environment when the environment does not provide information about the recommendation items in the dataset. In other words, the EWM sees limited semantic information, because the EWM prompt primarily represents videos through anonymous item and tag identifiers, which limits the usefulness of the LLM's semantic prior knowledge. This interpretation is consistent with the MovieLens anonymization results.}
    \label{fig:kuairand-anon}
\end{figure}

\end{document}